# *Alpha-N:* Shortest Path Finder Automated Delivery Robot with Obstacle Detection and Avoiding System

Asif Ahmed Neloy(✉)[0000-0002-2289-2762], Rafia Alif Bindu[0000-0002-6729-8548], Sazid Alam[0000-0001-8983-5124], Ridwanul Haque[0000-0002-4060-0436], Md. Saif Ahammod Khan[0000-0002-2666-8194], Nasim Mahmud Mishu[0000-0003-3973-0353], Shahnewaz Siddique

Department of Electrical and Computer Engineering, North South University, Plot-15, Block-B, Bashundhara Residential Area, Dhaka, Bangladesh
{asif.neloy,rafia.bindu,sazid.alam,ridwanul.haque,saif.ahammod,n asim.mishu,shahnewaz.siddique}@northsouth.edu

**Abstract.** *Alpha-N -* A self-powered, wheel-driven *Automated Delivery Robot (ADR)* is presented in this paper. The ADR is capable of navigating autonomously by detecting and avoiding objects or obstacles in its path. It uses a vector map of the path and calculates the shortest path by **Grid Count Method (GCM)** of Dijkstra's Algorithm. *Landmark determination with Radio Frequency Identification (RFID)* tags are placed in the path for identification and verification of source and destination, and also for the re-calibration of the current position. On the other hand, an **Object Detection Module (ODM)** is built by Faster R-CNN with VGGNet-16 architecture for supporting path planning by detecting and recognizing obstacles. The **Path Planning System (PPS)** is combined with the output of the GCM, the **RFID Reading System (RRS)** and also by the binary results of ODM. This PPS requires a minimum speed of 200 RPM and 75 seconds duration for the robot to successfully relocate its position by reading an RFID tag. In the result analysis phase, the ODM exhibits an accuracy of *83.75%*, RRS shows *92.3%* accuracy and the PPS maintains an accuracy of *85.3%*. Stacking all these 3 modules, the ADR is built, tested and validated which shows significant improvement in terms of performance and usability comparing with other service robots.

## 1    Introduction

Over the past few years, there has been an increasing number of robotic researches in Bangladesh. Especially a number of studies are conducted for Service Robots, Military Robots and also for the Rescue Robots. But it is rare to see robots in Bangladesh that are autonomous and capable of making complex decisions. An autonomous intelligent robot that can go from one place to another within a constructed map while avoiding an obstacle, is quite formidable to build. It comes with many challenges for solving this complex research topic. Moreover, the potential use of this kind of robot is in the food or parcel delivery systems which can be introduced in many rides sharing companies like Pathao Foods, Uber Eats, Food Panda in Bangladesh and also in developing countries [1]. Considering this huge opportunity and vacancy, this proposed ADR is studied.

This research particularly aims in the interaction between autonomous navigations using object detection. In such a problematic scenario, the robot has to detect and recognize objects as well as estimate their position by avoiding them. Although Object Detection and Recognizing are largely used in recent studies, most of them typically assume that the object is either already segmented from the background or that it occupies a large portion of the image. In ADR, to locate an object in any environment is very important because neither of the above predictions is correct since the distance to the object and also its size in the image can vary significantly. Therefore, the robot has to detect objects while moving with various frame rates and with the same trained object with different angles and sizes. This paper especially denotes to this problem and methods.

To module this complication, the proposed methodology uses Faster R-CNN having VGGNet-16 architecture for object detection that is especially suitable for detecting objects in a natural environment with complex systems [2], as it is able to cope with problems such as complex model, varying pixels and object perceptions. The construction of this type of system hasn't studied yet. Faster R-CNN uses selective search to find out the region proposals within an object. So, Faster R-CNN is much faster than its predecessors [2]. On the other hand, VGGNet-16 is a 16-layer model which shows the classification/localization accuracy within very little complex modeling. Therefore, it can be improved by increasing the depth of CNN in spite of using small receptive fields in the layers. For Alpha-N, a combination of both early learners and lower complexity with High Frame per Seconds (FPS) is considered for choosing the architecture. Faster R-CNN and VGG Net with 16 layers provide good results for this purpose which is illustrated in Section 5.

Due to the complex setup of the outdoor or even in the indoor environment, the path planning needs to recalibrate once every one cycle of planning is successfully completed. A possible solution for this recirculation is proposed where modeling approaches have been using as a cognitive map. Also, simultaneous localization and mapping (SLAM) and host localizations are proposed in existing studies [3,4]. But semantic and geometrical aspects are important for PPS with influential output from ODM. The ADR uses the vector map of the path between source and destination and calculates the shortest path using Dijkstra's Shortest Path Algorithm (GCM).

RFID Reader/Writer RC522 SPI S50 is considered for the operation of RRS. The output of the ODM and ODM controls the speed of the motor to scan and read the RFID tag. So, altogether Alpha-N combines the idea of an ADR having different object detection, avoiding, pathfinding algorithms to provide a solid fundamental improvement in the field of Industrial Robotics for Bangladesh.

## 2    Related Work

This paper extending the Previous work done by Neloy, A.A., et al. [5]. Previous work illustrates an Automated delivery robot made with RFID Scanner and Ultrasonic Sensor that can find its path from one point to another while avoiding obstacles, without any lines to follow. Although there exists a large number of works related to mobile robots with an object detection system or autonomous movement [6,7]. But there are still no fully operational systems that can operate robustly and long-term in indoor and outdoor environments combining obstacles detection and avoiding system. The current trend in the development of robots is divided onto single parts where a particular problem is studying. But the approach presented in this paper occupies both object detection along with autonomous navigation hand to hand.

There are some examples of such systems [8,9] where the robot is able to acquire and facilitate autonomous movement. Different to our methods, the works presented here [10], is a method of simultaneous localization and mapping (SLAM) to integrate both object detection and autonomous floor mapping. Along with SLAM, a Panoramic scan that automatically searches and generates the most precise and fast route which enables the robot to realize autonomous path planning path and so reaches the destination with the shortest route around obstacles [11]. A navigation system without a Grid map is also studied by [12]. These Autonomous robots are capable of move freely in a given circumstance, but most of them are not integrated with the idea of locating the distance between source and destination [13]. Also, previous work proposed by Tian, L et al. [14] isn't capable of adopting the idea of autonomous navigation by detecting obstacles from streaming or live video.

Most of the work on Dijkstra's algorithm resolves the pathfinding in a contained track. Also, the pathfinding problem exists in an environment where obstacles and constraints are situated statically. Kümmerle, R., et al. presented a navigation system for mobile robots that are designed to operate in crowded city environments and pedestrian zones [3]. They generated the local occupancy map by applying Dijkstra's algorithm to compute the distance to the nodes of the SLAM graph. A novel hierarchical global path planning approach for mobile robots in a cluttered environment is proposed by Mac, T. T., et al [4]. The authors' proposed Dijkstra's algorithm to find a collision-free path used as input reference for the next level. This study is very similar to ours but no module is offered to verify object location. Another study by Nazarahari, M., et al. proposed a hybrid approach for path planning of multiple mobile robots in continuous environments [15]. The proposed path length, smoothness, and safety are combined to form a multi-objective path planning problem. This study is also incomputable with autonomous movement in large complex systems.

The ADR we are presenting in this work needs to rank or navigate on the path based on the robust outliers of object recognition and avoid dynamic objects. No study is conducted where the robot both uses path planning consists of the shortest pathfinding and detecting, avoiding objects in that particular path. This paper bridges the gap between autonomous navigation accessing the shortest path passively observing the obstacles, viewing conditions and verifying by RFID tags in any dynamic environments.

Object recognition and detection have been extensively improved over a few years. Most of the approach is taken in deep learning techniques, among them, Faster R-CNN, Faster R-CNN, VGG Net are mostly studied which are region proposal based popular methods [2,16]. An object detection algorithm based on its color for the control of a mobile robotic platform with Android in a mobile robot is already studied by Gonzalez, E.M.A et al [17]. Vision techniques are introduced to produce this work. Holz, D., et al. propose a system for depalletizing and a complete pipeline for detecting and localizing objects as well as verifying that the acquired object does not deviate from the known object model [18]. They demonstrated the depalletizing of automotive and other prefabricated parts with both high reliability and efficiency. All of the states of art models symbolize the local method of object detection and recognizing. To do so, the object must appear large enough in the camera image or local features cannot be extracted from it. So, these models aren't capable to work on moving robots or may work with very little efficiency and accuracy. Moreover, the existing detector frameworks aren't built to generate any feedback to support path planning or localization. One of the contributions in this paper is the dynamic system that makes uses both approaches in a combined framework of path planning and detecting objects that let the methods complement each other.

The general idea of the overall system proposed in this paper is to learn action policies for the robot to locate a user-specified target destination from an initial point and reach the destination while avoiding obstacles by following an adequate algorithm. Also, it can identify the source and destination by RFID scanning. After delivering the items, the robot will backtrack its path to return on its source. Also, this extending study is improving both navigation, path analysis and can add a major contribution in both the hybrid navigation system constructed by Dijkstra's algorithm and avoiding obstacles by detecting objects. The outcome from both object detection and navigation provides precise performance for both indoor and outdoor environments.

## 3    Shortest Path Calculation by Vector Mapping and Dijkstra's Algorithm

The floor on which the Alpha-N will work on is first modeled into a graph. Each room is a node. Dijkstra's shortest path algorithm is modified to find the shortest distance between two nodes. First, Alpha-N needs a distance between source and destination. Suppose the Source is $U$ and the destination is $V$. The Alpha-N will calculate the shortest distance by Vector Mapping using the following algorithm [19]-

**Step 1.** The pathfinder introduces an Auxiliary Vector $D$, each of the component $D[i]$, the shortest path is discovered in the current starting point $U$, to each end $V_i$, then the $D[i]$ is the weight of the arc, otherwise $D[i] \rightarrow \infty$ and the length is-

$$D[i] = M_i^{in}\{D[i]|U_i \in V\} \tag{1}$$

This is the shortest path from $U$ to $V$ and stats as - $P(U_i, V)$.

**Step 2.** If the adjacency matrix arc with weights represent the directed graph and arc of the graph doesn't exist, then the arc[i][j] will be ∞. The initial value from map U to map V remains as $V_i$ and possible shortest path –

$$D[i] = arcs[Locate\ Vex(G, v)[i]\ ]\ vi \in V \tag{2}$$

**Step 3.** If the shortest path changes from the current $V_i$ to $V_K$, then the path will be updated as –

$$D[j] + arcs[j][k] < D[k] \tag{3}$$

Repeat until it reaches the initial state and updates the equation as –

$$D[k] = D[j] + arcs[j][k] \tag{4}$$

Using this updated $D[k]$ value, Dijkstra's Algorithm gets the weighted value of each node from source to destination. So, the value is denoting path cost for each decision. Shortest Path Grid Movement through Vector Map (Source to Destination and Destination to Source) is based on the 2-D grip. It defines the row and column between source and distance. So, this method calculates row to row and column to column distance to measure the actual path.

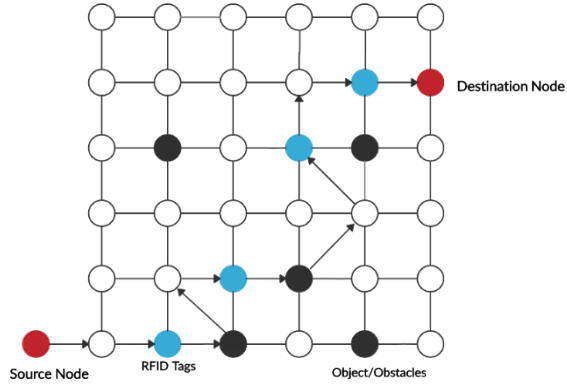

**Fig. 1.** A sample path following implementation using vector mapping and Grid Count from Equation 1 to 4.

## 4 Principles of the RFID Reading System (RRS)

The RFID tags can store information which supplies to any reader that is within a proximity range which can be up to approximately 15 meters. A special type of RFID Tag is used in this research which contains 6-digit. Based on the pattern of the Digit, the robot can make decisions. A brief classification of the tag digits is discussed in **Table 1**.

**Table 1.** RFID Tag Values

| Position ($xx - yy - zz$) | Values | | Decision |
|---|---|---|---|
| $xx$ | $xx$ values always 00 | | Initial or Source Node |
| $yy$ | 00 | 01 | The current Node is situated just after the Source Node, |
| | 01 | 10 | Middle Node. |
| | 10 | 11 | The Destination Node is after this current Node. |
| $zz$ | $zz$ values always 11 | | Destination Node. |

Based on the values, if the tag value is *00-01-11*, Alpha-N is in the source node, if *00-10-11* then Alpha-N is reached the destination and if 01-10-11 Alpha-N is in the path from source to destination.

## 5   Construction of ODM

The standard process of object detection and recognition for this research is consist of three steps-

- Detect the candidate regions of the object from the live streaming.
- Predict the class label of each region using bounding box regression and provide the final output.
- Denote each RPI to the next output for the path planning system.

However, there is two problem statement required for this task. The active perceiver setting for the models needs to detect specific objects for avoiding them. Following this observation, the authors adopt a deep neural network with leaser latency that simply takes account of the problem statements [20].

**Convolutional Networks and Region Proposal Network (RPN).** Based on the CAFFE framework [21], the output size of each convolutional and pooling layer can be calculated precisely by the following Equation 5 and 6)

$$output_{size} = \left\lfloor \frac{input_{size} + 2 \times pad - [dilation \times (kernel(k)_{size} - 1) + 1]}{stride} \right\rfloor + 1 \tag{5}$$

$$output_{pool} = \left\lceil \frac{input_{size} + 2 \times pad - kernel(k)_{size}}{stride} \right\rceil + 1 \tag{6}$$

where $\lfloor \rfloor$ and $\lceil \rceil$ are denoted as the floor and ceil function, respectively. For generating region proposals, each kernel location, $k = 9$, anchor boxes are used with 3 scales of 128, 256 and 512 values having 3 aspect ratios of 1: 1, 1: 2, 2: 1. Also, the area for each pooling area ($middle$) $= \frac{h}{H} \times \frac{w}{W}$ is obtained from the proposal [2]. Equation 5 and 6) provide the output area for each pooling 2×3 or 3×3 after rounding from the input Region of interest pooling (ROI) of 8×8. So, the output area for each pooling is 3×3 from the model.

**Classes and Bounding Boxes Prediction.** The final network predicts object class (classification) and Bounding boxes (Regression) from the model. This final output is also optimized by Stochastic Gradient Descent (SGD) which minimizes the convolution layers, RPN weights, and fully connected NN weights [2]. The depth of the feature map is 32 (9 anchors × 4 positions). Smooth-L1 loss function on the position $(x, y)$ having $top - left$ of the box (Equation 7 and 8) is proposed in this paper based on Equation 5 and Equation 6)

$$L_{loc}(t^u, v) = \sum_{i \in \{x,y,w,h\}} smooth_{L_1}(t_i^u - v_i) \tag{7}$$

$$\text{in which, } smooth_{L_1}(x) = \begin{cases} 0.5 \, x^2, \ if \ |x| < 1 \\ |x| - 0.5 \ otherwise, \end{cases} \tag{8}$$

The Classification and regression loss is the combination of the overall loss of RPN. The relation between this loss function and model accuracy is showed in **Fig. 3**.

## 5.1 Sample Dataset and Basic Setup

This paper uses the dataset of NYU V2 [22] for training and SUN RGB-D [23] for testing the model. The datasets are collected using an RGB-D camera in several indoor environments. A basic preprocessing and sample experimental setup are followed by previous work of van Beers, F. et al. [24]. A total number of 1400 of unique examples with 85 Classes are trained for the model. **Fig. 2** shows sample outputs from the model after the training phase. The output is primarily displayed or tested trough Pi camera.

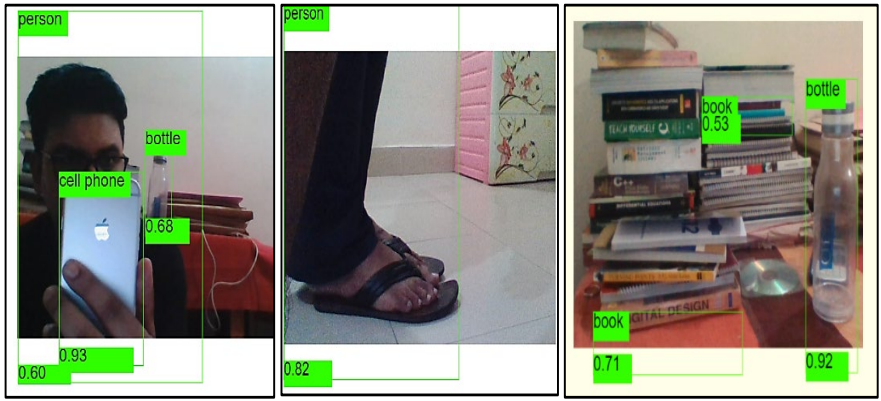

**Fig. 2.** Sample output of the model.

## 5.2 Training Parameter Setting and Accuracy

The pre-trained VGG16 model on ImageNet and Caffe Framework are applied during the training phase to initialize the parameters which are the forward channel of the network. The coefficient of the $smooth_{L_1} the$ loss function is set to 1, the learning ratio is set to 0.0075 after several experiments. **Fig. 3** shows the accuracy observation through training and testing curve. Overall accuracy is compared with two closely related papers. This score is recorded after test samples are fed to the model and the number of mistakes (zero-one loss) the model makes is accuracy. The problem of overfitting is minimized for this model by finding the optimum iterations with maximum accuracy and stops the epoch where when the accuracy is not improving, given a threshold. Previous work from Cho, S., et al [25] started this whole process.

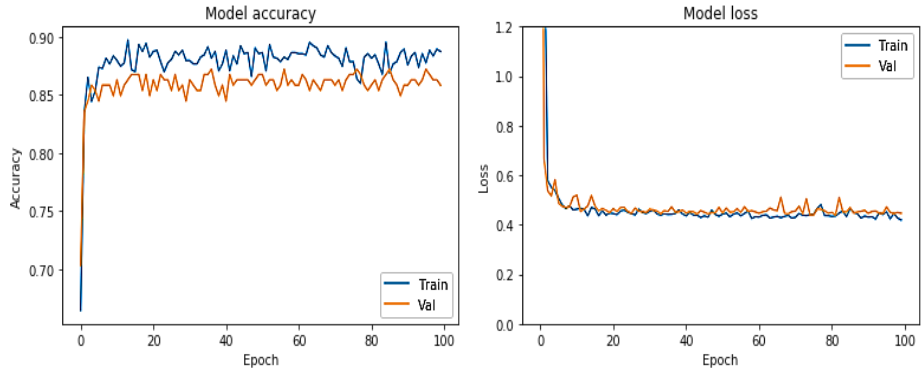

**Fig. 3.** Model Accuracy [left] and Model loss [right].

**Fig. 3**. exhibits early learner trends in the accuracy curve. The model could be trained more as the trend for accuracy on both curves is still rising for the last few epochs. Also, the model has not yet over-learned the training examples of boxplot, showing comparable skill on curve rising. Moreover, from the loss function graph, a relatively good comparable performance can be seen. the training might stop according to $smooth_{L_1}$ if it finds an overfitting score. Finally, the validation of the model is inspected by *Precision* and *Recall Curve* with metric proposed by Rezatofighi, H., et al[26]. **Table 2** and **Fig. 4**. show all the results derived from the validation process.

**Table 2.** 2-Fold cross-validation results of Faster R-CNN

| n-Fold | Recall | Precision | Avg Recall | Avg Precision | FP | FN |
|---|---|---|---|---|---|---|
| 1st Fold | 83.96 | 83.96 | 83.94 | 83.95 | 2110.2 | 2111.5 |
| 2nd Fold | 83.92 | 83.93 | | | | |

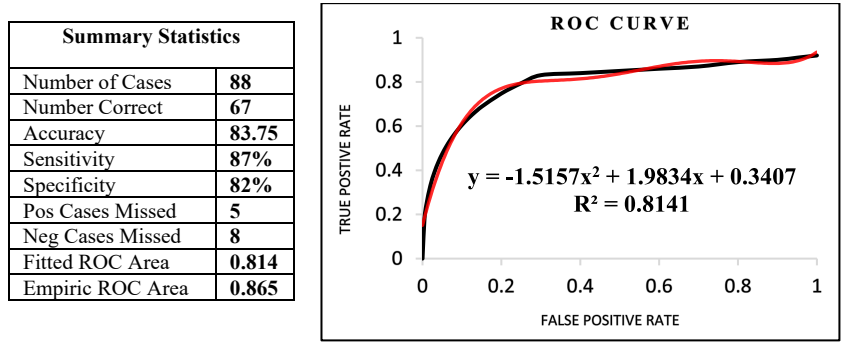

| Summary Statistics | |
|---|---|
| Number of Cases | **88** |
| Number Correct | **67** |
| Accuracy | **83.75** |
| Sensitivity | **87%** |
| Specificity | **82%** |
| Pos Cases Missed | **5** |
| Neg Cases Missed | **8** |
| Fitted ROC Area | **0.814** |
| Empiric ROC Area | **0.865** |

**Fig. 4.** ROC Curve Observation with Statistics.

The cross-validation result strongly connects the accuracy measurement from **Fig.** 3 Avg Accuracy obtained from the proposed model is *83.75%.* The overall result from the model is satisfactory compared to other works [2,20,26].

## 6    Testing the PPS

The shortest path algorithm implementation results using Equation 1,2,3,4 and the testing environment with sample-path and obstacles are presented in **Fig.** 5 [left] and path localization are presented in **Fig.** 5 [right]. After reaching the destination it waits for 120 seconds and backtracks to the origin or source.

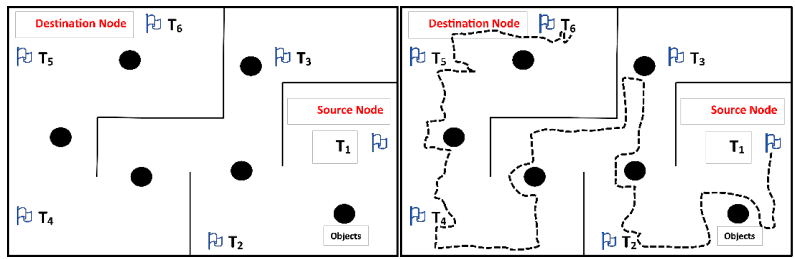

**Fig. 5.** The shortest path following by computing obstacle avoiding algorithm [right] is simulated in an indoor environment [left] with path and objects.

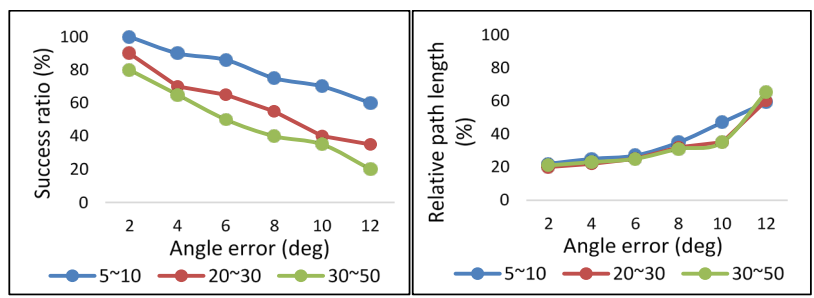

**Fig. 6.** 3 experiments constructed with 500 trials having a different number of RFID Tag numbers *(5~50)*. Success rate demonstrates in terms of angle [left] and path distance [right]. The figure shows the best performance ***(92.3%)*** with a *20~30* tag number and a relative path distance of *60* meters.

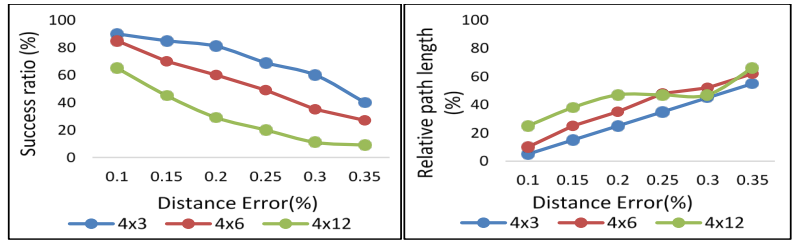

**Fig. 7.** 3 experiments constructed with 500 trials by varying the *grid number (4~12)*. Success rate demon-states in terms of traveled Distance. The figure shows the best performance ***(85.3%)*** with *4~3 grids* and a relative distance of *40 meters*.

## 7 CONCLUSION AND FUTURE WORK

This paper presents Alpha-N, an ADR with Shortest path Finder using Grid Count Algorithm, object or obstacle detecting module using Faster R-CNN with VGGNet-16, obstacle avoiding algorithm build with Improved Dynamic Window Approach(IDWA) and an Artificial Potential Field(APF) along with object detection. The primary objective of this research is to present a robot, which contributes to the research and industrial aspects of the robotics domain in Bangladesh and also in developing countries. It is shown that the overall experimental results of the system show satisfactory results and outcomes. Also, the results and working principle can be improved by examining the different environments and deploy on a large scale of entities.

In the future, all the frameworks will be tested in a hazardous environment with different scenarios to improve the sustainability of the system. More improvement of the design, framework, and modules will be studied as well.